\documentclass[final]{cvpr}

\usepackage{times}
\usepackage{epsfig}
\usepackage{graphicx}
\usepackage{amsmath}
\usepackage{amssymb}

\usepackage{colortbl}
\usepackage{multirow}
\usepackage{tablefootnote}
\usepackage{pifont}  
\usepackage{threeparttable}
\usepackage[numbers,sort]{natbib}
\usepackage{subcaption}
\usepackage{enumitem}
\usepackage[hang,flushmargin]{footmisc}

\setlist{nosep,leftmargin=1.2em,topsep=0.2em}

\definecolor{cmarkgreen}{RGB}{0,170,0}
\definecolor{xmarkred}{RGB}{225,0,0}

\newcommand{\paragraphheader}[1]{\vspace{4pt}\noindent\textbf{#1}\;}
\newcommand{\uptriangle}{\ding{115}}
\newcommand{\downtriangle}{\ding{116}}

\usepackage[pagebackref=true,breaklinks=true,letterpaper=true,colorlinks,bookmarks=false]{hyperref}

\def\cvprPaperID{****} 
\def\confYear{CVPR 2022}

\begin{document}

\title{The DEVIL is in the Details:\\ A Diagnostic Evaluation Benchmark for Video Inpainting}

\author{Ryan Szeto and Jason J. Corso\\
University of Michigan\\
{\tt\small \{szetor, jjcorso\}@umich.edu}
}

\maketitle

\begin{abstract}
   Quantitative evaluation has increased dramatically among recent video inpainting work, but the video and mask content used to gauge performance has received relatively little attention.
   Although attributes such as camera and background scene motion inherently change the difficulty of the task and affect methods differently, existing evaluation schemes fail to control for them, thereby providing minimal insight into inpainting failure modes.
   To address this gap, we propose the Diagnostic Evaluation of Video Inpainting on Landscapes (DEVIL) benchmark, which consists of two contributions: (i) a novel dataset of videos and masks labeled according to several key inpainting failure modes, and (ii) an evaluation scheme that samples slices of the dataset characterized by a fixed content attribute, and scores performance on each slice according to reconstruction, realism, and temporal consistency quality.
   By revealing systematic changes in performance induced by particular characteristics of the input content, our challenging benchmark enables more insightful analysis into video inpainting methods and serves as an invaluable diagnostic tool for the field. Our code and data are available at \href{https://github.com/MichiganCOG/devil}{github.com/MichiganCOG/devil}.
\end{abstract}

\vspace{-10pt}

\section{Introduction}

\begin{figure*}
  \includegraphics[width=\linewidth]{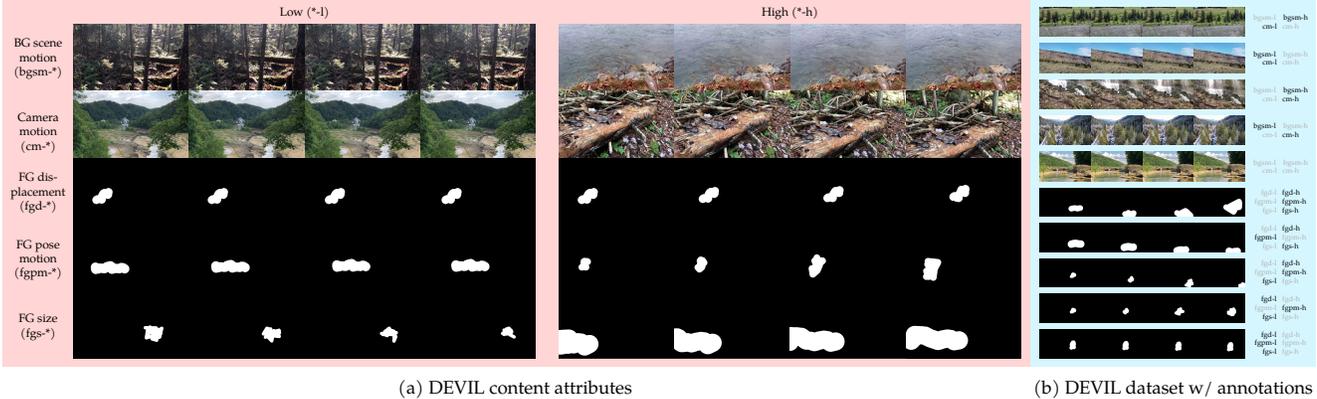}
  \caption{A visual overview of our DEVIL dataset. (a) The content attributes that characterize our dataset and are used to create dataset slices for evaluation (\ie, sets of video-mask pairs with a fixed attribute). We label low/high background scene motion or camera motion for videos exhibiting these attribute settings beyond a certain threshold (Section~\ref{sec:annotating-devil-source-video}). For occlusion masks, we construct sampling parameters that capture the desired attribute settings and use them to render masks (Section~\ref{sec:devil-masks-attributes}). (b) Videos, masks, and annotations from our dataset. A given video or mask may have multiple attribute labels or none; labels for the same attribute are mutually exclusive (\eg, a mask cannot have both low and high FG displacement).}
  \label{fig:dataset-overview}
  \vspace{-7pt}
\end{figure*}

Video inpainting, \ie, the task of filling in missing pixels in a video with plausible values, pushes the boundaries of modern video editing techniques and enables remarkable applications for film and social media such as watermark and foreground object removal~\cite{yeager_everything_2019,noauthor_production_2018}.
Compared to image inpainting, video inpainting is more challenging due to the additional temporal dimension, which not only increases the complexity of the solution space, but also places additional constraints on what constitutes a high-quality prediction---in particular, predictions must be coherent in terms of both spatial structure \textit{and} motion. Despite the difficulty of the task, modern results have become quite compelling thanks to the increasing amount of attention that the problem has received as of late~\cite{wang_video_2019,kim_deep_2019,xu_deep_2019,chang_free-form_2019,oh_onion-peel_2019,zhang_internal_2019,zeng_learning_2020,gao_flow-edge_2020}.

Quantitative evaluation has increased dramatically among recent video inpainting work; however, existing evaluation schemes underemphasize the importance of the contents of the videos and masks used to gauge performance.
Typically, video inpainting is evaluated as a reconstruction problem: performance is quantified by masking out arbitrary regions from the video (\ie, ``corrupting'' it) and scoring the model's ability to recover the masked-out values~\cite{xu_deep_2019,chang_free-form_2019,lee_copy-and-paste_2019}.
However, the difficulty of reconstruction depends on the mask's shape and motion, as well as the content present in the ``uncorrupted'' video.
For example, given a static mask, it is harder to inpaint a video captured by a fixed camera than one captured by a moving camera.
In the former case, the region beneath the mask is never visible, so the model effectively needs to ``hallucinate'' its appearance; in the latter case, the model could transfer appearance information from other frames, a strategy that lies at the heart of many video inpainting approaches~\cite{wexler_space-time_2007,huang_temporally_2016,lee_copy-and-paste_2019,oh_onion-peel_2019}.

\textit{The difficulty of video inpainting is inherently tied to the content of the videos and masks being inpainted;} with this principle in mind, we push for more emphasis on content-informed diagnostic evaluation, which can help identify the strengths and weaknesses of modern inpainting methods and improve ablative analysis.
To date, the videos used for evaluation have been underappreciated in this regard, having been sourced from datasets for other tasks (\eg, facial analysis~\cite{shen_first_2015,rossler_faceforensics_2018} and object localization~\cite{perazzi_benchmark_2016,xu_youtube-vos_2018}) rather than selected to represent important \textit{inpainting} scenarios.
In particular, they contain biases that are essential for the original task, but hinder fine-grained analysis for video inpainting.
For example, object localization videos consistently include prominent, moving foreground objects; as a result, standard inpainting evaluation schemes inevitably underrepresent performance on foreground-free videos.
Furthermore, other types of motion, \eg, camera and background scene motion, noticeably impact video inpainting performance, but are not controlled for in standard datasets.

In this work, we propose the Diagnostic Evaluation of Video Inpainting on Landscapes (DEVIL) benchmark.
It is composed of two parts---the DEVIL dataset and the DEVIL evaluation scheme---which combine to enable a finer-grained analysis than has been possible in prior work.
Such granularity is achieved through \textit{content attributes}, \ie, properties of source videos or masks that characterize key failure modes by affecting how easily video inpainting models can borrow appearance information from nearby frames.
Specifically, the DEVIL dataset contains source videos labeled with low/high camera and background scene motion attributes, and occlusion masks labeled with low/high foreground displacement, pose motion, and size attributes (Figure~\ref{fig:dataset-overview}).
Meanwhile, the DEVIL evaluation scheme constructs several \textit{slices} of the DEVIL dataset---sets of video-mask pairs in which exactly one content attribute is kept fixed---and summarizes inpainting quality through metrics that capture reconstruction performance, realism, and temporal consistency (Section~\ref{sec:evaluation-metrics}).
By controlling for content attributes and summarizing inpainting quality per attribute across several metrics, our DEVIL benchmark provides valuable insight into the failure modes of a given inpainting model and how mistakes manifest in the output.

We use our novel benchmark to analyze the strengths and weaknesses of seven state-of-the-art video inpainting methods.
By quantifying their inpainting quality on ten DEVIL dataset slices under five evaluation metrics, we provide the most comprehensive and fine-grained evaluation of modern video inpainting methods to our knowledge.
Our head-to-head, multi-faceted comparisons allow us to draw several important conclusions.
For example, we show that video inpainting methods in which time and optical flow are carefully modeled consistently achieve the best performance across several types of input data.
We also show that the relative rankings between methods are highly sensitive to metrics as well as source video and mask content, highlighting the need for comprehensive evaluation.
Finally, we show that controlling for source video and mask attributes reveals insightful failure modes which can be traced back to the design of the inpainting method in question.

Our comprehensive diagnostic benchmark enables insightful analysis and serves as an invaluable tool for video inpainting research. To summarize, we provide the following contributions:
\begin{itemize}
  \item We present the first diagnostic dataset specifically designed for video inpainting to our knowledge, which includes annotations for content-based attributes that represent numerous inpainting failure modes;
  \item We introduce a novel and comprehensive evaluation scheme that spans ten dataset slices and five evaluation metrics of video inpainting quality;
  \item We analyze seven state-of-the-art algorithms on our benchmark, providing the most comprehensive quantitative evaluation of video inpainting methods to date; and
  \item We identify systematic errors among video inpainting methods and highlight directions for future work.
\end{itemize}
Our benchmark is available at \url{https://github.com/MichiganCOG/devil}.

\section{Related Work}

\subsection{Methods}
\label{sec:methods}

Most video inpainting algorithms borrow visual appearance information from known parts of the video to fill in unknown parts. For example, \textit{alignment-based methods} compute local or global alignments between neighboring frames, and then propagate pixels across aligned locations (alignments are found either through classical feature correspondences~\cite{granados_background_2012,ebdelli_video_2015} or deep neural networks~\cite{lee_copy-and-paste_2019,oh_onion-peel_2019,zeng_learning_2020}). \textit{Patch-borrowing methods} view videos as spatiotemporal tensors and iteratively paste cuboids or voxels from the known region into the unknown region to maximize global coherence~\cite{wexler_space-time_2007,granados_how_2012,newson_video_2014}. \textit{Flow-guided methods} propagate visual information along the optical flows estimated between consecutive frames, where the flow for unknown regions is computed iteratively or hierarchically based on known regions to improve performance~\cite{huang_temporally_2016,kim_deep_2019,xu_deep_2019,gao_flow-edge_2020}.

Although explicit appearance-borrowing is not leveraged for some methods, \eg, \textit{autoencoder-based methods}~\cite{wang_video_2019,chang_free-form_2019,chang_learnable_2019,zhang_internal_2019}, we use its prevalence to guide our experimental design. Specifically, we control for content-based attributes that affect the difficulty of borrowing relevant appearance information from other frames, which is possible through our varied data and comprehensive annotations.

\subsection{Datasets}
\label{sec:datasets}

The source video datasets used in early video inpainting work were small and oriented around qualitative analysis~\cite{wexler_space-time_2007,granados_how_2012,newson_video_2014}; as a result, it was difficult to compare the performance of early methods across a wide variety of scenarios. Recent video inpainting methods have instead used large-scale datasets---ranging in structure from aligned face videos~\cite{rossler_faceforensics_2018} and driving videos~\cite{dollar_pedestrian_2012} to unconstrained videos from the Internet~\cite{perazzi_benchmark_2016,xu_youtube-vos_2018}---to enable more comprehensive analysis. The foreground object segmentation datasets DAVIS~\cite{perazzi_benchmark_2016} and YouTube-VOS~\cite{xu_youtube-vos_2018} have been particularly popular since their annotations can be used to remove the object from the video. Unlike most prior work, we collect novel videos specifically for the inpainting task, emphasizing attributes that affect how well video inpainting models can transfer appearance information across frames. We also use videos of background scenes instead of videos with foreground objects, which enables us to isolate failure modes that are unrelated to foreground motion.

In terms of occlusion masks, static rectangles~\cite{xu_deep_2019,wang_video_2019} or foreground masks from other videos~\cite{kim_deep_2019,lee_copy-and-paste_2019,oh_onion-peel_2019,zhang_internal_2019} are commonly used. Procedural mask generation has also been explored; for instance, Chang \etal~\cite{chang_learnable_2019,chang_free-form_2019} render several strokes on a canvas, where each stroke has several control points that randomly move with a certain probability. We extend their work by adding physical constraints for finer control over mask size and motion. Furthermore, we choose our mask sampling parameters based on attributes that directly influence how easily video inpainting models can transfer appearance information across frames.

\section{Overview of the DEVIL Benchmark}

Before introducing the DEVIL benchmark for video inpainting, we first define the task itself. Let $V \in \{0, \dots, 255\}^{H \times W \times 3 \times T}$ be an input RGB video with $T$ frames and a resolution of $W \times H$. $V$ contains a placeholder value for missing voxels (\eg, 0) whose locations are indicated by an input occlusion mask $M \in \{0, 1\}^{H \times W \times T}$.
Video inpainting aims to produce an inpainted version of $V$, denoted $V^*$, with the following characteristics:
\begin{itemize}
  \item \textbf{Reconstruction performance:} $V^*$ is a faithful reconstruction of $V^{\textrm{gt}}$ in the scenario where $V$ is a ``corrupted'' video derived from some uncorrupted ground truth source video $V^{\textrm{gt}}$.
  \item \textbf{Realism:} $V^*$ is indistinguishable from a real video.
  \item \textbf{Temporal consistency:} $V^*$ exhibits minimal temporal flickering artifacts.
\end{itemize}
These criteria are defined more rigorously in Section~\ref{sec:evaluation-metrics}.

Our DEVIL benchmark is a collection of tools designed to provide a detailed understanding of video inpainting methods and their behavior across a variety of input data. There are two major components of our benchmark: (i) the DEVIL dataset, which contains source videos and occlusion masks that have been specially curated, rendered, and annotated to identify specific failure modes in video inpainting; and (ii) the DEVIL evaluation scheme, which reports a set of quality-based metrics on several ``slices'' of the DEVIL dataset, each of which represents a particular failure mode.

The DEVIL dataset captures content complexity along five video-level \textit{content attributes}, \ie, properties that affect the difficulty of inpainting a given video-mask pair by influencing the relevance and availability of appearance information from nearby frames. Specifically, it contains source videos with low and high camera and background (BG) scene motion, and occlusion masks with low and high foreground (FG) displacement, pose motion, and size (Figure~\ref{fig:dataset-overview}a).\footnote{The term ``foreground'' (FG) comes from FG object removal applications.} Furthermore, videos and masks are annotated with these attributes and their settings (low or high) to enable targeted evaluation that controls for their presence. Section~\ref{sec:devil-dataset} rigorously defines these attributes and describes our process for collecting videos, masks, and attribute annotations.

Meanwhile, our DEVIL evaluation scheme gauges inpainting quality under multiple \textit{slices} of our dataset, \ie, sets of video-mask pairs characterized by a certain attribute setting. Within each slice, exactly one dataset attribute is kept fixed while the others change freely. By measuring inpainting quality across several slices and metrics, our benchmark provides valuable information on when and how models fail. In Section~\ref{sec:devil-evaluation}, we describe our DEVIL dataset slices and evaluation metrics in further detail.

\section{The DEVIL Dataset}
\label{sec:devil-dataset}

\subsection{Collecting Source Videos for the DEVIL}
\label{sec:collecting-source-videos-devil}

In the context of quantitative evaluation, video inpainting is generally posed as a reconstruction problem~\cite{xu_deep_2019,chang_free-form_2019,lee_copy-and-paste_2019}; thus, it is useful to evaluate on videos of background scenes without foreground objects, where the complete ground-truth background appearance is known (and foreground behavior can be controlled explicitly via occlusion masks). Data from other video understanding tasks do not satisfy this criterion, since they generally feature foreground objects which ground the original task. This is especially true for the two most popular datasets used in video inpainting work, DAVIS~\cite{perazzi_benchmark_2016} and YouTube-VOS~\cite{xu_youtube-vos_2018}, which were originally collected for foreground object segmentation.

For this reason, we collect our own videos of background-only scenes, similar to Zhang \etal~\cite{zhang_internal_2019}. In particular, we target scenic landscape videos in which people have filmed natural outdoor locations from both casual and cinematic viewpoints. Because the primary subject of these videos is the background, they are substantially less likely to contain prominent foreground objects, and are thus good targets for curating our source video collection.

To collect scenic landscape videos, we first search Flickr~\cite{smugmug_inc_flickr_nodate} using the query term ``scenic'', and retain videos from a fixed set of users who have primarily uploaded high-quality, non-post-processed content between 2017-2019. Then, we apply a combination of automated and manual filtering to remove videos with foreground objects or shot transitions. Finally, we split the filtered videos into clips containing between 45-90 frames, constituting a total of 1,250 clips (examples are shown in Figure~\ref{fig:dataset-overview}). Additional details are provided in the supplementary materials.

\subsection{Annotating DEVIL Source Video Attributes}
\label{sec:annotating-devil-source-video}

For our DEVIL source videos, we annotate two types of content attributes: camera motion and BG scene motion (Figure~\ref{fig:dataset-overview}a). \textbf{Camera motion} encompasses frame-to-frame differences that are induced by changes in the camera's pose relative to the scene (\ie, camera extrinsics); \textbf{BG scene motion} refers to frame-to-frame differences that result from changes in the scene itself, such as running bodies of water or trees that sway due to strong winds (\ie, motion among ``stuff'' classes in the object detection sense~\cite{forsyth_finding_1996}).

We select these attributes for two reasons. First, they represent two sources of complex motion with different low-level characteristics that video inpainting models must replicate well to produce convincing predictions. Second, they impact video inpainting models by influencing the similarity and relevance of appearance information across frames. For instance, high camera motion can reveal or obscure parts of the scene, or otherwise change the scene's appearance due to perspective; high BG scene motion continuously changes the frame-wise appearance of textures.

These attributes are difficult to quantify concretely based on RGB video frames alone; however, it is possible to distinguish extreme examples of low and high motion by visual inspection and proxy estimates. Thus, for a given attribute, we label videos as containing either low or high motion, but only for a small percentage of videos lying at the extreme ends (videos not labeled for the given attribute may still appear in slices that do not control for it). Not only does this reduce label ambiguity, it also magnifies any performance differences caused by changing a given attribute between low and high settings, thereby highlighting failure modes.

To annotate camera motion, we use classical affine alignment techniques and measure the amount of invalid pixels introduced via warping as a proxy for camera motion; we then threshold the result on either side to produce low and high motion labels. As for BG scene motion, we manually assign low and high labels based on the percentage of the field of view that contains large running bodies of water. Further details are provided in the supplementary materials.

\subsection{DEVIL Masks and Attributes}
\label{sec:devil-masks-attributes}

For occlusion masks, we consider three attributes that influence the availability of relevant appearance information in nearby frames (Figure~\ref{fig:dataset-overview}a):
\begin{itemize}
  \item \textbf{FG displacement:} How much the mask's centroid moves over time with respect to the field of view;
  \item \textbf{FG pose motion:} How much the shape of the mask changes over time with respect to the field of view (independent of the displacement of its centroid); and
  \item \textbf{FG size:} The average number of pixels that are occupied by the mask per frame.
\end{itemize}
Masks with high FG displacement or pose motion reveal complementary parts of the scene over time, whereas FG size explicitly determines how much appearance information can be relied on as ground truth.

To generate masks with low and high settings for these attributes, we adapt the procedural blob generation strategy by Chang \etal~\cite{chang_free-form_2019}. In particular, we adjust their stroke width, velocity, and stochasticity parameters to correspond to low or high FG displacement, pose motion, and size. For more details on the mask generation process, including our extensions to enable finer control over individual blobs, refer to the supplementary materials.

\section{The DEVIL Evaluation}
\label{sec:devil-evaluation}

\subsection{Slices of the DEVIL Dataset}

The na\"ive way to evaluate on the DEVIL dataset would be to randomly sample a test set of paired source videos and occlusion masks without accounting for their attributes; however, this provides little insight into the failure modes that cause prediction errors for a given method. Instead, we control for one attribute at a time to isolate its impact on the prediction. Specifically, for each attribute setting, we construct \textit{slices} of the DEVIL dataset, \ie, pre-determined sets of video-mask pairs where the given attribute is fixed as low or high and the others are uncontrolled. By reporting performance on each slice separately, our benchmark can highlight failure modes with finer granularity.

We construct the DEVIL dataset slices as follows. Given the desired attribute setting (\eg, low camera motion), we randomly sample either 150 source videos or masks with that setting (recall that an attribute applies exclusively to either the source video or the mask modality). Then, within the other modality, we sample 150 instances from all available DEVIL instances (\eg, for the low camera motion slice, we sample all rendered DEVIL masks). Finally, we pair together the selected source videos and masks.

\subsection{Evaluation Metrics}
\label{sec:evaluation-metrics}

\begin{figure}
  \centering
  \begin{subfigure}{0.3\linewidth}
    \begin{subfigure}{\linewidth}
      \includegraphics[width=\linewidth]{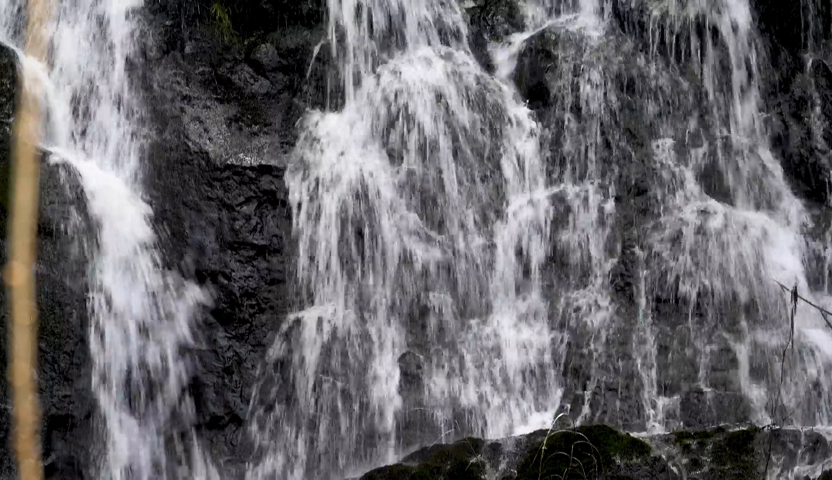}
    \end{subfigure}
    \begin{subfigure}{\linewidth}
      \includegraphics[width=\linewidth]{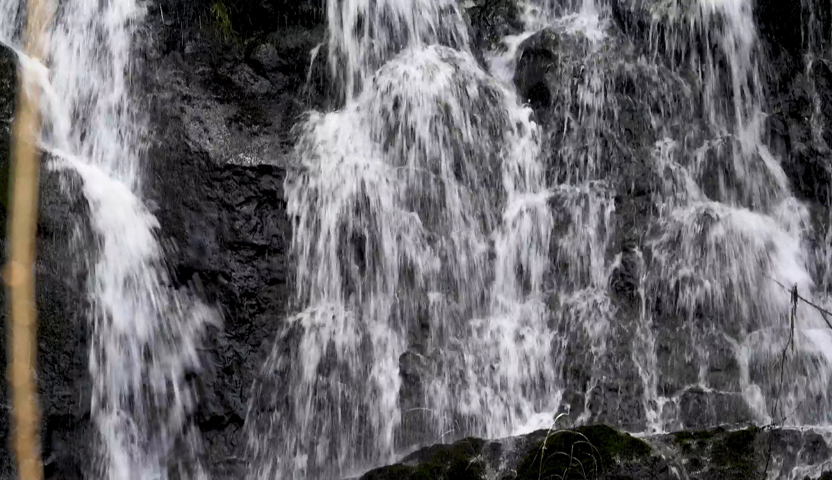}
    \end{subfigure}
    \caption{Ground truth}
  \end{subfigure}
  \hspace{1pt}
  \begin{subfigure}{0.3\linewidth}
    \begin{subfigure}{\linewidth}
      \includegraphics[width=\linewidth]{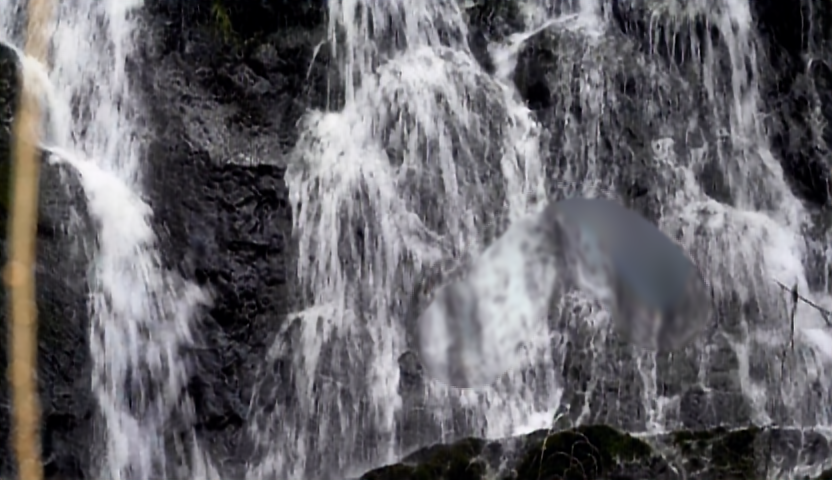}
    \end{subfigure}
    \begin{subfigure}{\linewidth}
      \includegraphics[width=\linewidth]{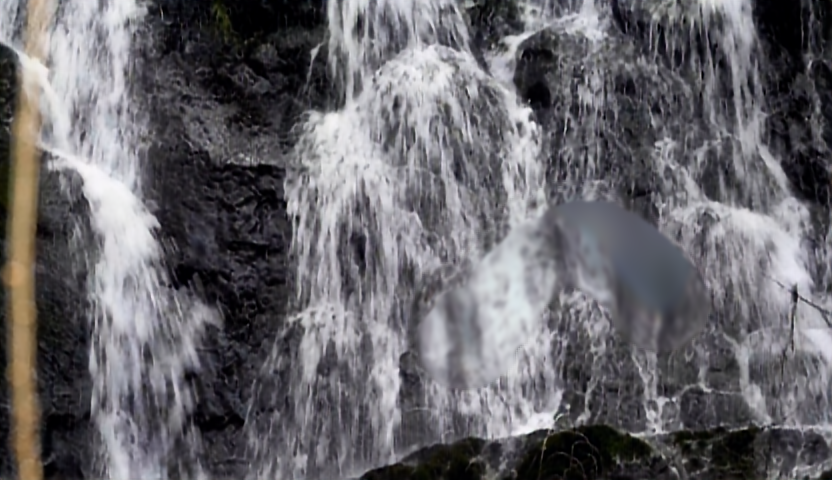}
    \end{subfigure}
    \caption{VINet}
  \end{subfigure}
  \hspace{1pt}
  \begin{subfigure}{0.3\linewidth}
    \begin{subfigure}{\linewidth}
      \includegraphics[width=\linewidth]{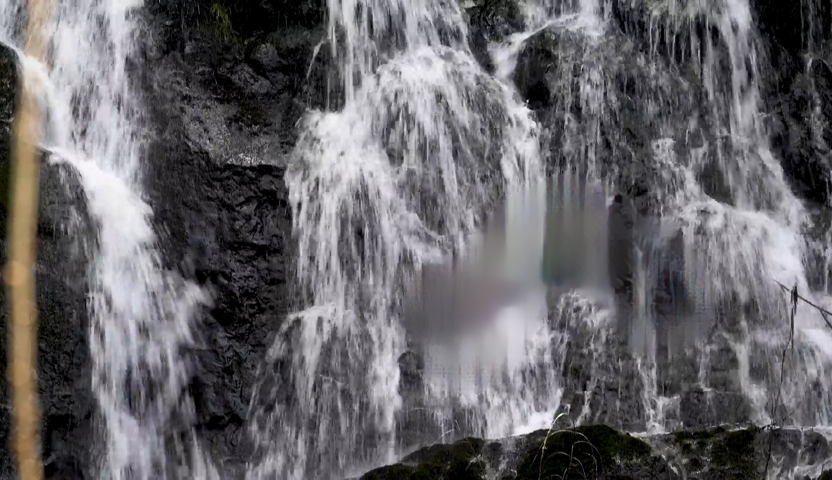}
    \end{subfigure}
    \begin{subfigure}{\linewidth}
      \includegraphics[width=\linewidth]{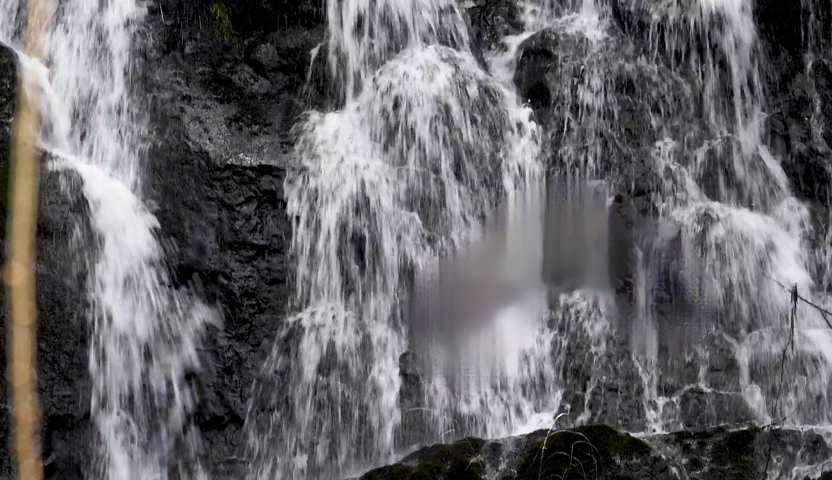}
    \end{subfigure}
    \caption{CPNet}
  \end{subfigure}
  \caption{High temporal consistency may indicate overly blurry predictions if reconstruction or realism performance is low (as shown in the results of VINet and CPNet). The area to be inpainted is outlined in yellow in (a).}
  \label{fig:blurry}
  \vspace{-10pt}
\end{figure}

We evaluate on composited inpainting results (\ie, the known region is composited over the inpainting prediction). Inputs are resized to 832$\times$480 resolution; for methods that cannot consume this resolution, we apply mirror padding to both the source video and the mask, run the method, and crop out padded regions from the result for fairness. We quantify performance along three axes of inpainting quality---reconstruction, realism, and temporal consistency---as described in the remainder of this section.

\paragraphheader{Reconstruction} captures the extent to which a video inpainting method predicts the original content in a given reference video (\ie, the version of the video without the occlusion mask). We report two reconstruction metrics: the Learned Perceptual Image Patch Similarity (LPIPS) metric~\cite{zhang_unreasonable_2018} with a pre-trained AlexNet backbone~\cite{krizhevsky_imagenet_2012}, and our own video-based variant called the Perceptual Video Clip Similarity (PVCS) metric with a pre-trained I3D backbone~\cite{carreira_quo_2017}. These metrics measure the distance between corresponding features of a deep neural network. LPIPS is computed between corresponding frames of the reference and inpainted video, whereas PVCS is computed between corresponding 10-frame clips in a sliding window.

\paragraphheader{Realism} indicates how well the inpainting result resembles the appearance and motion observed in a reference set of real videos, independent of the original video from which the input is derived. Unlike reconstruction, realism enforces a smaller penalty for deviating from the original content as long as the prediction exhibits sensible appearance and motion.
For our image-based realism metric, we use the Fr\'echet Inception Distance~\cite{heusel_gans_2017} (FID), which fits multivariate normal distributions over the feature activations of two sets of images and measures their distance; in our case, the two sets correspond to all predicted frames and all reference frames. We also report the video-based equivalent Video FID~\cite{kim_deep_2019} (VFID), which corresponds to the sets of all inpainted videos and all reference videos (the features are extracted from a pre-trained I3D backbone~\cite{carreira_quo_2017}).

\paragraphheader{Temporal consistency} measures the proliferation of flickering artifacts, \ie, how much colors at corresponding points of the scene change between consecutive frames. We adapt the patch consistency metric, denoted PCons, from Gupta \etal~\cite{gupta_characterizing_2017}: for each frame, we extract the 50$\times$50 patch at the centroid of the occlusion mask, compute the maximum Peak Signal-to-Noise Ratio (PSNR) between this patch and neighboring patches in the next frame, and average the result across all frames. Note that stronger temporal consistency is not always ideal: low-quality predictions, such as constant-color or blurry inpainting results, can produce high temporal consistency scores (see Figure~\ref{fig:blurry}).

\section{Experiments}
\label{sec:experiments}

\begin{figure*}
  \begin{subfigure}[b]{0.72\linewidth}
    \includegraphics[width=\linewidth]{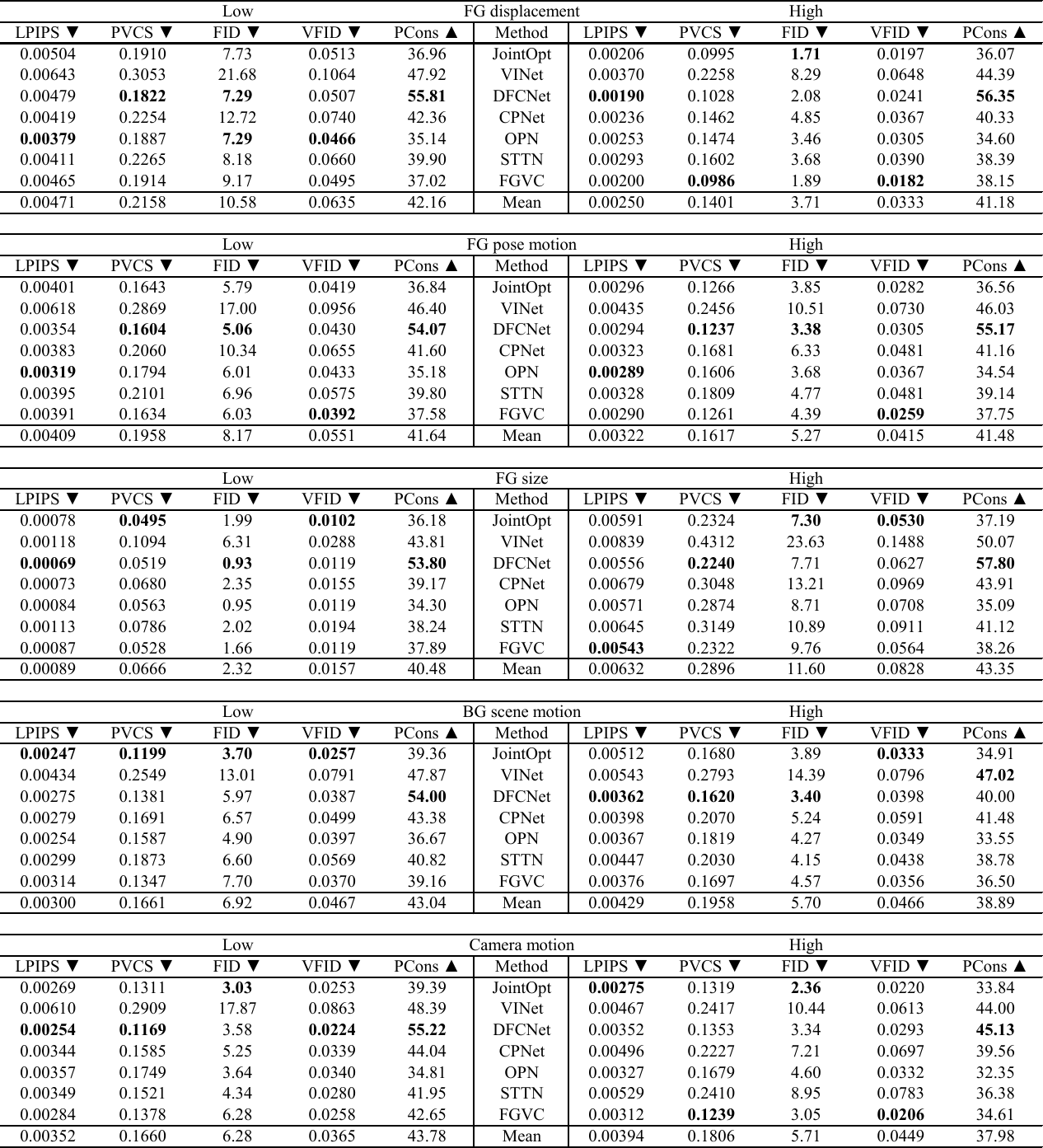}
    \stepcounter{table}
    \vspace{-10pt}
    \caption*{Table \thetable: The performance of each inpainting method on each DEVIL slice and evaluation metric. Bold indicates the best method; \downtriangle{} and \uptriangle{} indicate that lower or higher is better, respectively.}
    \label{tab:absolute-performance}
  \end{subfigure}
  \hfill
  \begin{subfigure}[b]{0.25\linewidth}
    \begin{subfigure}{0.98\linewidth}
      \includegraphics[trim=0px 30px 0px 30px,clip,width=\linewidth]{figs/model-performance/lpips}
    \end{subfigure}
    \begin{subfigure}{0.98\linewidth}
      \includegraphics[trim=0px 30px 0px 30px,clip,width=\linewidth]{figs/model-performance/pvcs}
    \end{subfigure}
    \begin{subfigure}{0.98\linewidth}
      \includegraphics[trim=0px 30px 0px 30px,clip,width=\linewidth]{figs/model-performance/fid}
    \end{subfigure}
    \begin{subfigure}{0.98\linewidth}
      \includegraphics[trim=0px 30px 0px 30px,clip,width=\linewidth]{figs/model-performance/vfid}
    \end{subfigure}
    \begin{subfigure}{0.98\linewidth}
      \includegraphics[trim=0px 30px 0px 30px,clip,width=\linewidth]{figs/model-performance/pcons}
    \end{subfigure}
    \caption*{Figure \thefigure: Mean performance of each method across all DEVIL slices. Error bars show standard error across the ten slices.}
    \label{fig:model-performance}
  \end{subfigure}
  \vspace{-7pt}
\end{figure*}

To demonstrate the utility of our DEVIL benchmark, we analyze the performance of seven representative state-of-the-art video inpainting methods, using the publicly-available code, model weights, and default runtime arguments provided by the original authors:
\begin{itemize}
  \item \textbf{Joint optimization of flow and color (JointOpt)~\cite{huang_temporally_2016}:} Alternates between optimizing an optical flow estimate and finding suitable patches along the flow. Among our methods, this is the only non-deep learning one.
  \item \textbf{VINet~\cite{kim_deep_2019}:} Recurrently predicts the next frame by warping intermediate features spatially via optical flow.
  \item \textbf{Deep Flow Completion Network (DFCNet)~\cite{xu_deep_2019}:} Predicts the optical flow of the video, then inpaints the missing region by propagating known values along the flow.
  \item \textbf{Copy-Paste Network (CPNet)~\cite{lee_copy-and-paste_2019}:} Estimates affine transformations between frames with a task-driven deep neural network, and then copies features across aligned frames via attention.
  \item \textbf{Onion Peel Network (OPN)~\cite{oh_onion-peel_2019}:} Iteratively inpaints the exterior of the current missing region by attending to relevant locations in the known region.
  \item \textbf{Spatio-Temporal Transformer Network (STTN)~\cite{zeng_learning_2020}:} Decodes the missing region with a Transformer~\cite{vaswani_attention_2017} that consumes multi-scale patches from the entire video.
  \item \textbf{Flow-Edge Guided Video Completion (FGVC)~\cite{gao_flow-edge_2020}:} Extends DFCNet~\cite{xu_deep_2019} by leveraging flow between non-adjacent frames and using edge information to solve for piecewise-smooth flow predictions.
\end{itemize}

\subsection{Aggregate Analysis}

In Table~\hyperref[tab:absolute-performance]{1}, we report the performance of all evaluated methods on each DEVIL slice; in Figure~\hyperref[fig:model-performance]{3}, we compare their performance averaged across all slices. We observe that JointOpt, DFCNet, and FGVC consistently outrank or perform within one standard error of the other methods in terms of the reconstruction metrics LPIPS/PVCS and the realism metrics FID/VFID. They all explicitly solve for the optical flow of the inpainted video during inference, suggesting that \textit{computing task-driven flow is an essential ingredient in producing the highest-quality video inpainting results}. Additionally, JointOpt remains competitive among recent deep learning-based solutions, suggesting that improvements may arise by adapting traditional subroutines, \eg, PatchMatch~\cite{barnes_patchmatch_2009}, to deep learning.

The three mid-tier methods---OPN, STTN, and CPNet---borrow intermediate features across distant time steps, but do not use time as an ordered structure. Meanwhile, VINet models time through a recurrent unit, but has the shortest temporal receptive field among the evaluated methods and cannot propagate information from future time steps back through the entire video. These results indicate that \textit{modeling time as a proper ordered structure with long-range dependencies greatly improves inpainting quality.}

In terms of temporal consistency, DFCNet achieves the highest PCons since it propagates pixel values directly along the predicted flow maps of adjacent frames. Interestingly, JointOpt and FGVC achieve lower PCons despite also propagating values along the flow, likely due to their ability to transfer candidate values from non-adjacent frames. VINet and CPNet achieve high temporal consistency at the cost of low-quality predictions lacking appropriate texture or motion (Figure~\ref{fig:blurry}); the behaviors of these two models indicate that temporal consistency is most meaningful when reconstruction and realism performance is also high.

\begin{figure}
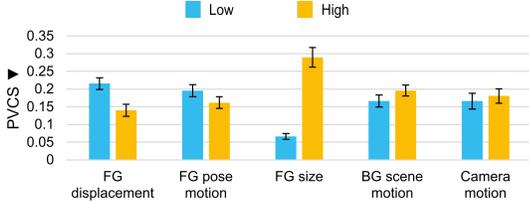

  \begin{subfigure}{\linewidth}
    \centering
    \includegraphics[width=0.9\linewidth]{figs/slice-difficulty/legend}
  \end{subfigure}
  \begin{subfigure}{\linewidth}
    \centering
    \includegraphics[trim=0px 20px 0px 0px,clip,width=0.9\linewidth]{figs/slice-difficulty/pvcs}
  \end{subfigure}
  \caption{Comparison of DEVIL slice difficulty based on average model performance (lower is better; error bars show standard error across the seven models). The type of content given at test time, especially the mask content, greatly affects the difficulty of the task.}
  \label{fig:slice-difficulty}
\end{figure}

\begin{table}
  \centering
  \includegraphics[width=0.75\linewidth]{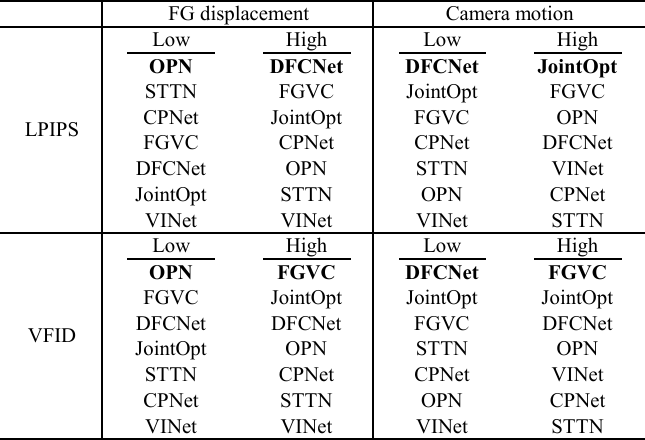}
  \caption{Methods sorted by performance from best to worst based on three variables: the metric (LPIPS/VFID), the \mbox{attribute} (FG displacement/camera motion), and the setting of said attribute (low/high). The strongest method (highlighted in bold) depends on all three variables, showing that no one method dominates our challenging benchmark.}
  \label{fig:rank}
\end{table}

\paragraphheader{DEVIL Attribute Difficulty} We now analyze how DEVIL attributes impact overall inpainting difficulty to highlight their utility in video inpainting evaluation. In Figure~\ref{fig:slice-difficulty}, we compare the difficulty of each DEVIL attribute setting by computing the average PVCS over all methods on the corresponding test slice. The mask attributes substantially impact the difficulty of video inpainting; in particular, higher FG displacement and pose motion, as well as smaller FG size, lead to better performance. These trends make sense---more relevant appearance information is available in other frames when the occlusion mask is smaller and moves more over time. In contrast, the overall impact of camera and BG scene motion is small due to the differing sensitivities of each method to these attributes (Section~\ref{sec:model-sensitivity-devil-attributes}). We observe similar trends under the other reconstruction and realism metrics (refer to the supplementary materials for details).

\subsection{Model Sensitivity to DEVIL Attributes}
\label{sec:model-sensitivity-devil-attributes}

\begin{figure}
  \begin{subfigure}{\linewidth}
    \centering
    \includegraphics[width=0.86\linewidth]{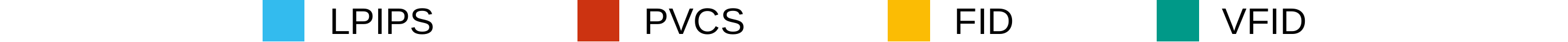}
  \end{subfigure}
  \begin{subfigure}{\linewidth}
    \centering
    \includegraphics[width=0.9\linewidth]{figs/relative-difference/cm}
  \end{subfigure}
  \caption{Relative improvement of each method under reconstruction and realism metrics when camera motion changes from low to high.}
  \label{fig:relative-difference-cm}
  \vspace{-2pt}
\end{figure}

\begin{figure}
  \centering
  \begin{subfigure}{0.4\linewidth}
    \centering
    \begin{subfigure}{0.95\linewidth}
      \includegraphics[width=\linewidth]{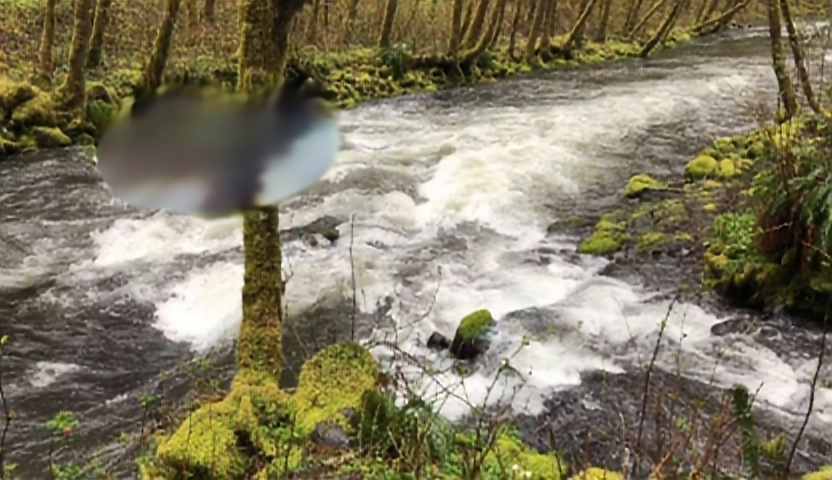}
    \end{subfigure}
    \begin{subfigure}{0.95\linewidth}
      \includegraphics[width=\linewidth]{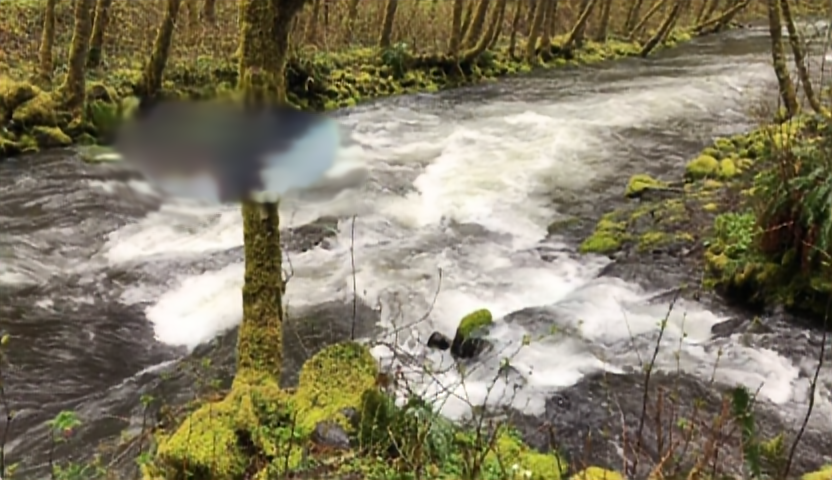}
    \end{subfigure}
    \caption{Low camera motion}
  \end{subfigure}
  \begin{subfigure}{0.4\linewidth}
    \centering
    \begin{subfigure}{0.95\linewidth}
      \includegraphics[width=\linewidth]{figs/vinet-cm/cm-h/vinet/frame_0000_pred}
    \end{subfigure}
    \begin{subfigure}{0.95\linewidth}
      \includegraphics[width=\linewidth]{figs/vinet-cm/cm-h/vinet/frame_0020_pred}
    \end{subfigure}
    \caption{High camera motion}
  \end{subfigure}
  \caption{VINet prediction examples. VINet improves with high camera motion since content ``moves into'' the missing region (arrows show the direction of camera motion).}
  \label{fig:vinet-cm}
  \vspace{-10pt}
\end{figure}

Next, we compare the performance of each method on individual DEVIL slices to demonstrate the impact of video and mask content, as well as the metric, on their rankings. Table~\ref{fig:rank} lists the methods sorted by performance and grouped by three variables: (i) the metric, (ii) the DEVIL attribute, and (iii) the particular setting of the attribute (\ie, low or high). Note that the two attributes shown, FG displacement and camera motion, span the two modalities of the DEVIL dataset (masks and videos, respectively); also note that these three variables span the complexity of the DEVIL evaluation scheme in terms of metrics and dataset slices. Even among the eight different combinations of variables shown out of the possible 50, four methods achieve the best performance for at least one combination; this demonstrates that our DEVIL benchmark is a substantial challenge for video inpainting methods.

Finally, we analyze the sensitivity of the methods to each DEVIL attribute by changing the label and plotting the relative difference in performance. Specifically, for a given attribute, we compute the relative improvement as $(\textrm{score}_\textrm{hi} - \textrm{score}_\textrm{lo}) / \textrm{score}_\textrm{lo}$ (or the negation of this value if a lower score is better), where $\textrm{score}_\textrm{hi}$ and $\textrm{score}_\textrm{lo}$ correspond to model performance on the high and low slices of the attribute, respectively. As evidenced by the following observations, our DEVIL attributes enable a fine-grained comparison of failure modes among different inpainting models.

In Figure~\ref{fig:relative-difference-cm}, we see divergent behavior among the methods when camera motion is increased: for example, STTN and CPNet experience dramatic performance drops under all reconstruction and realism metrics, whereas VINet experiences substantial gains. From a design standpoint, the behaviors of STTN and CPNet make sense because they rely on aligning patches or entire frames from other time steps to borrow their features, which can fail catastrophically under heavy camera motion. On the opposite end, VINet's behavior reflects its tendency to inpaint realistic textures only after they move into the missing portion of the frame (Figure~\ref{fig:vinet-cm}), which is unlikely to occur without camera motion.

\begin{figure}
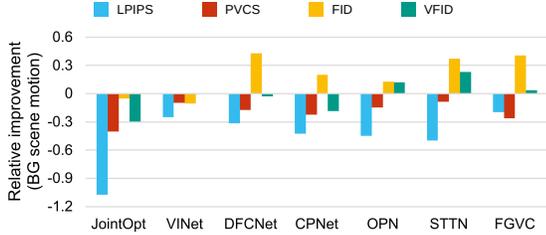

  \begin{subfigure}{\linewidth}
    \centering
    \includegraphics[width=0.86\linewidth]{figs/relative-difference/legend-video}
  \end{subfigure}
  \begin{subfigure}{\linewidth}
    \centering
    \includegraphics[width=0.9\linewidth]{figs/relative-difference/bsm}
  \end{subfigure}
  \caption{Relative improvement of each method under reconstruction and realism metrics when BG scene motion changes from low to high.}
  \label{fig:relative-difference-bsm}
\end{figure}

\begin{figure}
  \centering
  \begin{subfigure}{0.24\linewidth}
    \begin{subfigure}{\linewidth}
      \includegraphics[width=\linewidth]{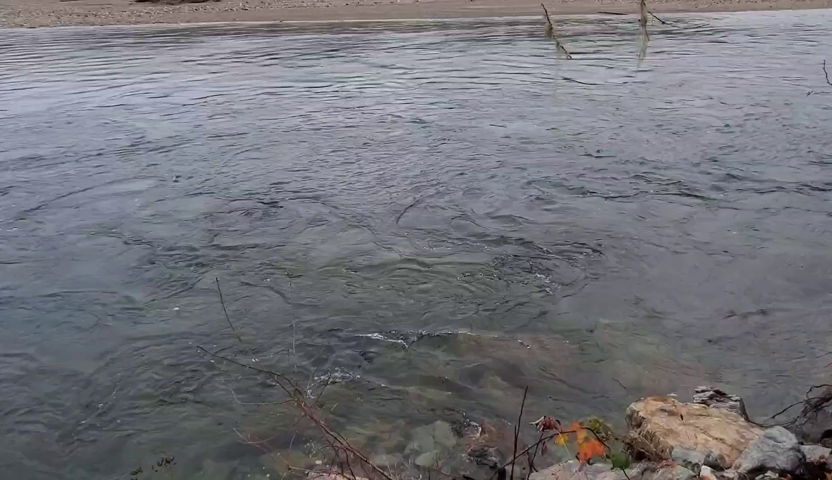}
    \end{subfigure}
    \begin{subfigure}{\linewidth}
      \includegraphics[width=\linewidth]{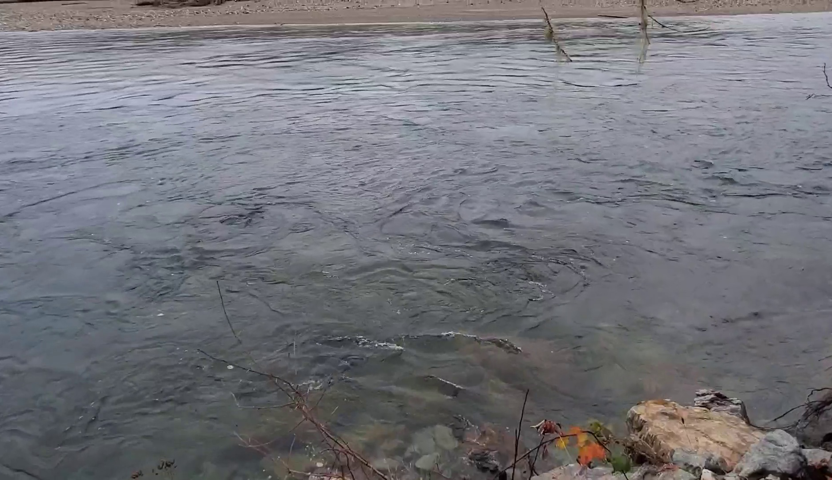}
    \end{subfigure}
    \caption{GT}
  \end{subfigure}
  \begin{subfigure}{0.24\linewidth}
    \begin{subfigure}{\linewidth}
      \includegraphics[width=\linewidth]{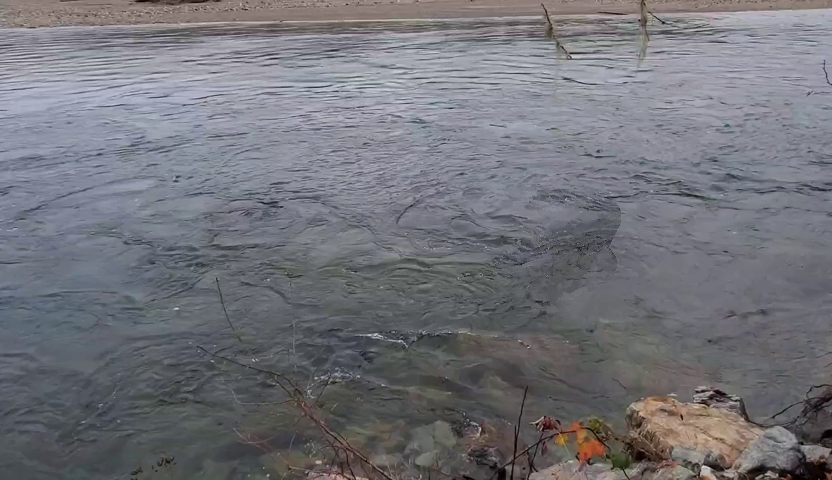}
    \end{subfigure}
    \begin{subfigure}{\linewidth}
      \includegraphics[width=\linewidth]{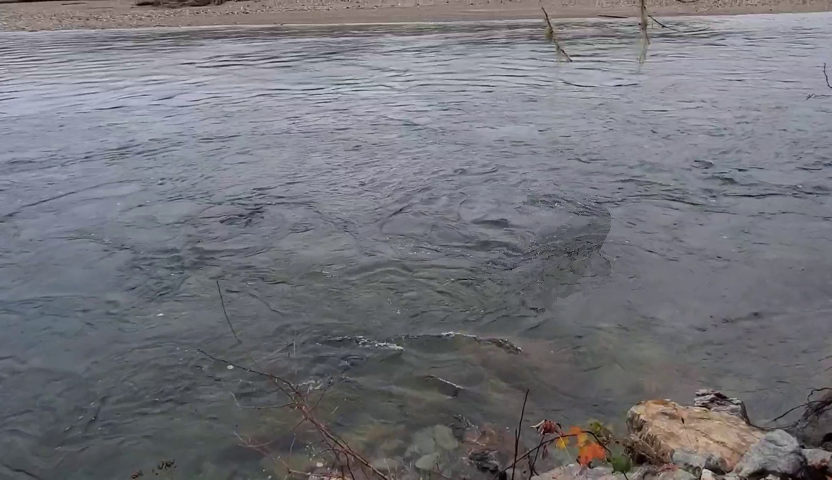}
    \end{subfigure}
    \caption{DFCNet}
  \end{subfigure}
  \begin{subfigure}{0.24\linewidth}
    \begin{subfigure}{\linewidth}
      \includegraphics[width=\linewidth]{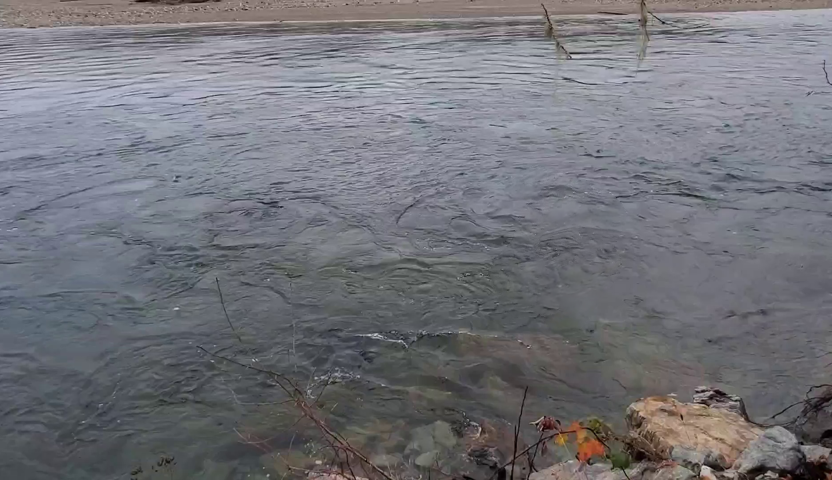}
    \end{subfigure}
    \begin{subfigure}{\linewidth}
      \includegraphics[width=\linewidth]{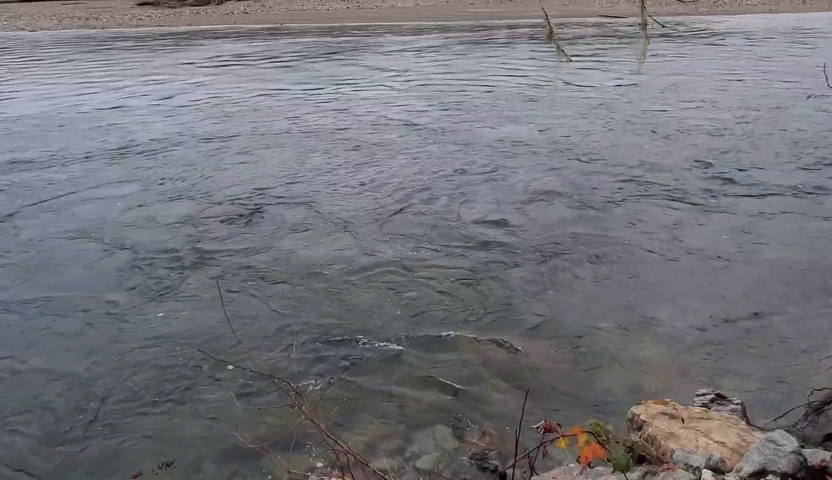}
    \end{subfigure}
    \caption{FGVC}
  \end{subfigure}
  \caption{Example inpainting predictions from the high BG scene motion slice. For DFCNet and FGVC, the predictions diverge from the original content, but still exhibit semantically sensible appearance.}
  \label{fig:alt-reality}
  \vspace{-7pt}
\end{figure}

\begin{figure}
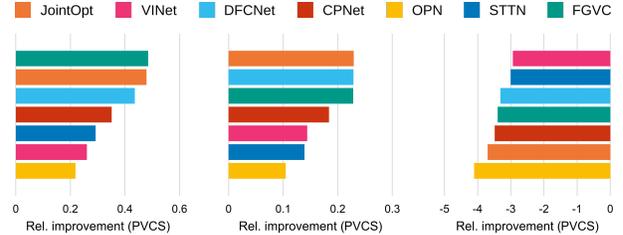

  \begin{subfigure}{\linewidth}
    \centering
    \includegraphics[width=0.95\linewidth]{figs/relative-difference/legend-mask}
  \end{subfigure}
  \begin{subfigure}{\linewidth}
    \begin{subfigure}{0.32\linewidth}
      \includegraphics[width=\linewidth]{figs/relative-difference/fgd}
      \caption{FG displacement}
    \end{subfigure}
    \hfill
    \begin{subfigure}{0.32\linewidth}
      \includegraphics[width=\linewidth]{figs/relative-difference/fgm}
      \caption{FG pose motion}
    \end{subfigure}
    \hfill
    \begin{subfigure}{0.32\linewidth}
      \includegraphics[width=\linewidth]{figs/relative-difference/fgs}
      \caption{FG size}
      \label{fig:relative-difference-mask-fgs}
    \end{subfigure}
  \end{subfigure}
  \caption{Relative improvement of each method under PVCS when mask attributes change from low to high.}
  \label{fig:relative-difference-mask}
  \vspace{-5pt}
\end{figure}

From Figure~\ref{fig:relative-difference-bsm}, we see that reconstruction performance under LPIPS and PVCS consistently worsens when BG scene motion increases, reflecting the challenge of replicating complex dynamics precisely. Interestingly, frame realism under FID actually improves dramatically for some methods such as DFCNet and FGVC because they inpaint ``alternate realities'', \ie, appearance that diverges from the original content, but still captures the original broad structure (Figure~\ref{fig:alt-reality}). Among the methods that explicitly infer and propagate across flow (FGVC, JointOpt, and DFCNet), only JointOpt achieves worse frame realism performance, suggesting that \textit{video inpainting quality can be improved by simply learning flow in a task-driven, end-to-end manner.}

Moving on to mask attributes, we show in Figure~\ref{fig:relative-difference-mask}a-b that all methods perform better with increased FG displacement and pose motion, indicating that they all leverage the increased availability of background information. The flow propagation methods generally benefit the most, likely due to the explicit transmission of more ground-truth appearance information along predicted flows. We observe similar trends under the other reconstruction and realism metrics (details are available in the supplementary materials).

In Figure~\ref{fig:relative-difference-mask-fgs}, we observe worse performance when the mask FG size grows, reflecting the inherently greater difficulty of inpainting more values. Although this trend is universal and intuitive, the quantitative difference between methods still lends insight into their failure modes. For example, OPN is most sensitive to the increased mask size because it must iteratively inpaint more outer layers of the unknown region based on its own predictions, thereby accumulating error across more iterations.

\section{Conclusion}

We have presented the DEVIL benchmark for video inpainting and used it to analyze seven state-of-the-art methods, thereby providing the largest fine-grained analysis of video inpainting to our knowledge. By controlling for five content attributes of the source videos and masks used at test time, our analyses have provided novel insight into the behaviors, strengths, and weaknesses of these methods.


\section*{Acknowledgements}

We appreciate the exploratory contributions made by Xuetong Sun and Jing-An Tzeng. This material is based upon work supported by the National Science Foundation under Grant No.~1628987. Any opinions, findings, and conclusions or recommendations expressed in this material are those of the author(s) and do not necessarily reflect the views of the National Science Foundation.

\newpage
\onecolumn

\vspace*{0.5cm}
\begin{center}
  \Large \textbf{The DEVIL is in the Details:\\ A Diagnostic Evaluation Benchmark for Video Inpainting\\Supplementary Materials}
\end{center}
\vspace*{1.5cm}

\appendix
\section{DEVIL Dataset Details}

\subsection{Source Videos}

To identify a high-quality set of source videos depicting scenic landscapes, we begin by searching Flickr~\cite{smugmug_inc_flickr_nodate} for videos that contain the term ``scenic'' in their metadata. From these preliminary results, we identify a small number of users who upload a large volume of high-quality, non-post-processed videos. We then refine our search to ``scenic'' videos from those users within a given upload time frame (January 2017 - January 2019). From these videos, we automatically detect and discard any that contain shot transitions, resolutions not equal to 1920$\times$1080, or COCO object classes~\cite{lin_microsoft_2014} as detected by a Mask R-CNN model~\cite{he_mask_2017} provided by Detectron2~\cite{wu_detectron2_2019}. After automatic filtering, we manually inspect the remaining videos and remove those that contain undetected foreground objects or shot transitions, as well as other signs of post-processing (\eg, sped-up videos). We split the remaining videos into clips containing between 45-90 frames, which constitute a grand total of 1,250 source clips.

\subsection{Source Video Attributes}

To annotate high BG scene motion, we manually identify clips that contain running bodies of water that cover at least 40\% of the frame for all frames; for low BG scene motion, we identify clips that contain no running bodies of water (we establish our BG scene motion annotations based on bodies of water since they are prevalent in our data and easy for people to identify by visual inspection). We did not use automatic classifiers for this attribute due to their poor performance and the automatic bias that would have been introduced through their usage.

To annotate camera motion, we use classical affine alignment techniques and measure the amount of invalid pixels introduced via warping as a proxy for camera motion. The intuition behind this classifier is that high camera motion produces frames with poor pairwise affine alignments, and that warping frames by such transforms introduces a high percentage of invalid pixels into the field of view (the converse is true for low camera motion). Despite the simplicity of this approach, we found that it achieves a sufficiently high precision-recall AUC for our purposes (0.90 on a manually-annotated version of the DAVIS train/val set~\cite{perazzi_benchmark_2016}).

Concretely, we label camera motion as follows: between a given pair of video frames, we first compute bidirectional robust affine transformations using RANSAC~\cite{fischler_random_1981} over matched SURF keypoints~\cite{bay_surf_2006}. Then, we warp the frames by the corresponding affine transformation and compute the number of invalid pixels introduced by the warp; we define the inverse of this quantity as the pairwise compatibility between the given frames. For a given clip, we sample ten evenly-spaced frames and compute the minimum pairwise compatibility between all pairs, which we define as the total frame compatibility of the clip. Finally, we obtain camera motion annotations by thresholding the total frame compatibility.

\subsection{Occlusion Masks}

To generate occlusion masks with our desired DEVIL attributes, we opt for a procedural generation approach inspired by Chang \etal~\cite{chang_free-form_2019}, which enables fine-grained control over mask shape and behavior. In their framework, an initial mask shape is generated by sampling control points along a random walk with momentum (\ie, biased toward an initial direction), and then connecting the control points with a stroke of random thickness. The mask is animated by moving all control points with a given velocity and then slightly perturbing their positions at each time step.

We extend the code of Chang \etal with several changes to enable even finer control over mask size and motion. For example, we reduce the impact of momentum in the initial mask-drawing phase to increase the diversity of mask shapes. Additionally, we apply inward-facing acceleration to the control points whenever they are sufficiently far from the mask's centroid, which effectively constrains its maximum possible area. Furthermore, we force the control points to bounce off the edge of the frame to prevent them from leaving the field of view. Finally, we randomly reverse the temporal dimension of masks with a 50\% probability since they tend to grow in size over time.

Because the mask generation procedure is parameterized, we can produce occlusion masks that correspond to our desired DEVIL attribute settings by sampling from distinct configurations. To generate masks with small and large FG sizes, we sample from two corresponding ranges of stroke widths, and also change the maximum possible distance of each control point to the centroid. To vary the FG displacement, we sample the initial velocity of the overall mask from two different ranges. Finally, to generate masks with low and high pose motion, we vary the stochasticity of the control points (\ie, for low pose motion masks, the control points are less likely to accelerate in a random direction per frame).

\section{Evaluation Metric Details}

\begin{figure}
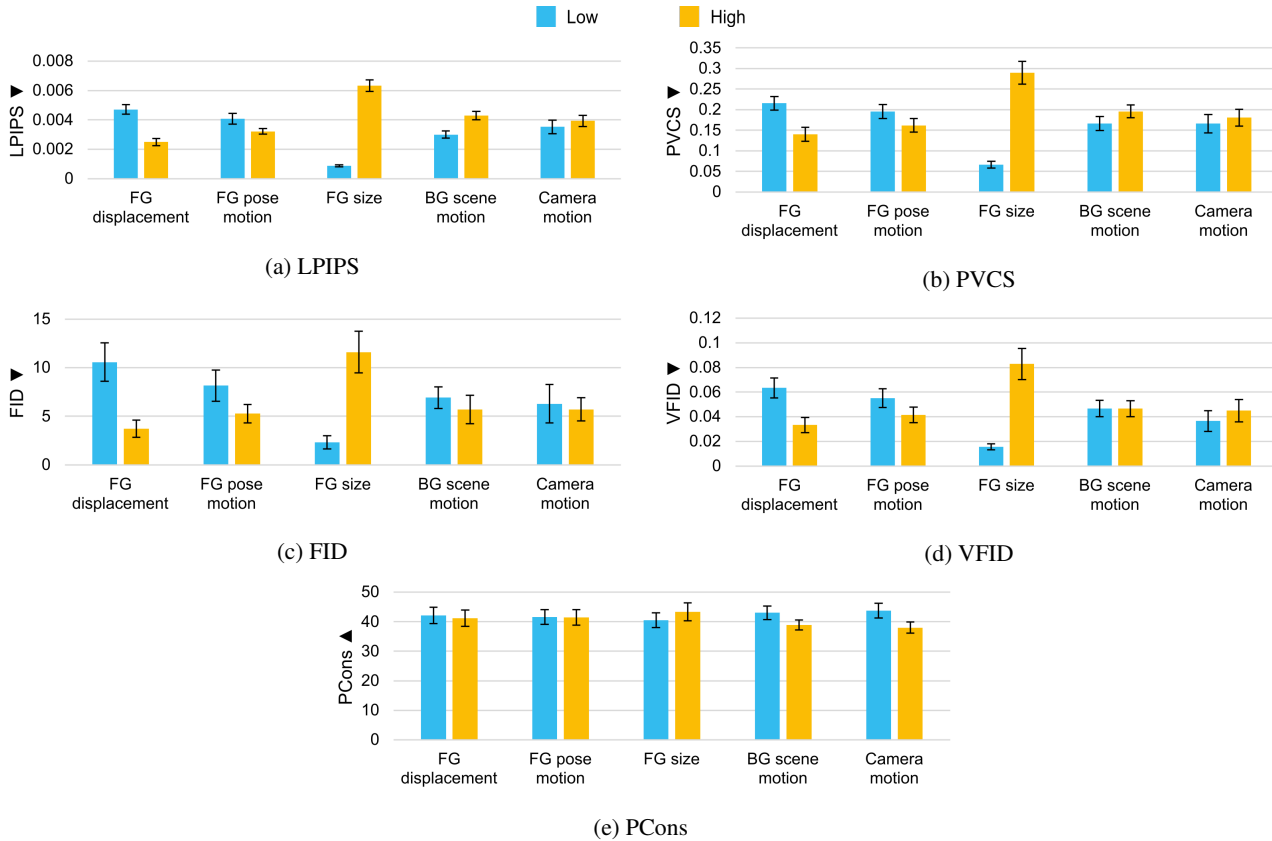

  \centering
  \begin{subfigure}{\linewidth}
    \centering
    \includegraphics[width=0.55\linewidth]{figs/slice-difficulty/legend}
  \end{subfigure}
  \begin{subfigure}{0.5\linewidth}
    \includegraphics[width=\linewidth]{figs/slice-difficulty/lpips}
    \caption{LPIPS}
  \end{subfigure}%
  \begin{subfigure}{0.5\linewidth}
    \includegraphics[width=\linewidth]{figs/slice-difficulty/pvcs}
    \caption{PVCS}
  \end{subfigure}
  \begin{subfigure}{0.5\linewidth}
    \includegraphics[width=\linewidth]{figs/slice-difficulty/fid}
    \caption{FID}
  \end{subfigure}%
  \begin{subfigure}{0.5\linewidth}
    \includegraphics[width=\linewidth]{figs/slice-difficulty/vfid}
    \caption{VFID}
  \end{subfigure}
  \begin{subfigure}{0.5\linewidth}
    \includegraphics[width=\linewidth]{figs/slice-difficulty/pcons}
    \caption{PCons}
  \end{subfigure}
  \caption{Comparison of DEVIL slice difficulty. \downtriangle{} and \uptriangle{} indicate that lower and higher is better, respectively. Error bars show standard error across the seven evaluated methods.}
  \label{fig:supp-slice-difficulty}
\end{figure}

In this section, we further describe our evaluation metrics, including details on the features used for the deep neural network-based metrics and parameters for the temporal consistency metric.

\paragraphheader{LPIPS and PVCS}
Our implementation of LPIPS is derived from the original code from Zhang \etal~\cite{zhang_unreasonable_2018}. We use their fine-tuned AlexNet model weights~\cite{krizhevsky_imagenet_2012} as well as their feature activations. For PVCS, we extend the LPIPS code to use a pre-trained I3D model~\cite{carreira_quo_2017} in place of the AlexNet model; distance is computed from the feature activations from I3D's five pre-pooling layers.

\paragraphheader{FID and VFID}
Our implementation of FID is derived from a third-party implementation of the metric from Heusel \etal~\cite{heusel_gans_2017}.\footnote{The third-party implementation is available at \url{https://github.com/mseitzer/pytorch-fid}.} The representation of a video frame corresponds to the activations from the final pooling layer of the Inception Network~\cite{szegedy_going_2015}, followed by global mean pooling over the remaining spatial dimensions.
For VFID, we extend the third-party implementation of FID to use the same pre-trained I3D model as PVCS. To obtain the representation of a video, we extract the activations of I3D's final pooling layer and compute the average over the spatial and temporal dimensions (VFID is thus the Fr\'echet distance over video representations).

\paragraphheader{PCons}
To compute PCons between two frames, we first extract the 50$\times$50 patch centered at the centroid of the mask from the first frame (if this patch partially lies beyond the boundary, we clip the centroid coordinate such that the patch lies entirely inside the image). Then, we compute the maximum PSNR between the extracted patch and all valid 50$\times$50 patches in the second frame whose centers are within a Chebyshev distance of at most 20 pixels from the first frame's centroid coordinate. To compute the PCons of an entire video, we take the average PCons over all consecutive frame pairs (\ie, a 2-frame sliding window).

\section{Additional Quantitative Results}

In Figure~\ref{fig:supp-slice-difficulty}, we show the average performance of the seven evaluated inpainting methods on each of our ten DEVIL splits. We observe that across the reconstruction and realism metrics (Figure~\ref{fig:supp-slice-difficulty}a-d), performance changes substantially under the occlusion mask attributes, but less substantially under source video attributes. The temporal consistency metric PCons changes less dramatically under the DEVIL attributes (Figure~\ref{fig:supp-slice-difficulty}e), suggesting that temporal consistency performance is relatively stable under changes in the source video and mask content.

\begin{figure}
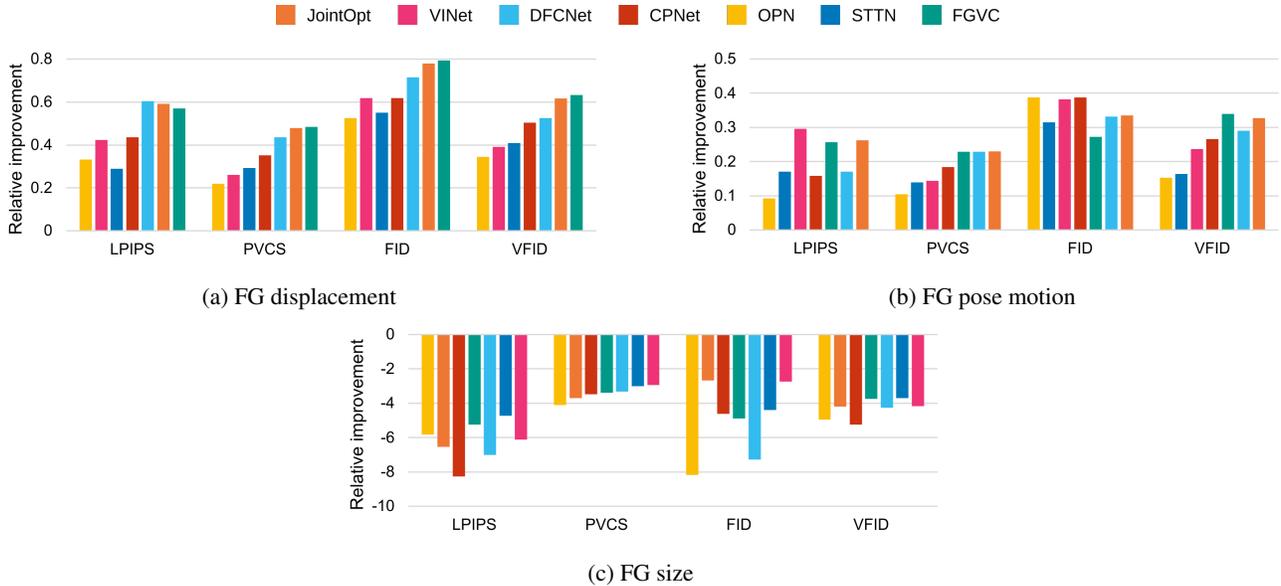

  \centering
  \begin{subfigure}{\linewidth}
    \centering
    \includegraphics[width=0.55\linewidth]{figs/relative-difference/legend-mask}
    \vspace{5pt}
  \end{subfigure}
  \begin{subfigure}{\linewidth}
    \begin{subfigure}{0.48\linewidth}
      \includegraphics[width=\linewidth]{figs/supp-relative-difference/fgd}
      \caption{FG displacement}
    \end{subfigure}
    \hfill
    \begin{subfigure}{0.48\linewidth}
      \includegraphics[width=\linewidth]{figs/supp-relative-difference/fgm}
      \caption{FG pose motion}
    \end{subfigure}
  \end{subfigure}
  \begin{subfigure}{0.48\linewidth}
    \includegraphics[width=\linewidth]{figs/supp-relative-difference/fgs}
    \caption{FG size}
  \end{subfigure}
  \caption{Relative improvement of each method under reconstruction and realism metrics when DEVIL mask attributes change from low to high. Within each plot, the methods are sorted by PVCS performance.}
  \label{fig:supp-relative-difference}
\end{figure}

In Figure~\ref{fig:supp-relative-difference}, we show the relative improvement experienced by each method when FG displacement, pose motion, and size are increased. The flow propagation methods FGVC, JointOpt, and DFCNet generally benefit the most from increased FG displacement and pose motion, whereas OPN benefits the least. As for FG size, OPN is more sensitive to this attribute than the other methods under three out of four reconstruction and realism performance metrics.

\begin{figure}
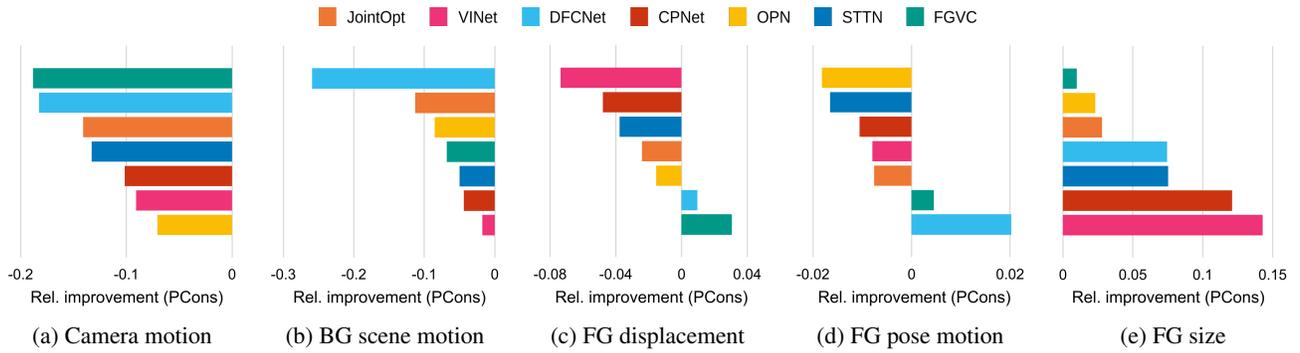

  \centering
  \begin{subfigure}{\linewidth}
    \centering
    \includegraphics[width=0.5\linewidth]{figs/relative-difference/legend-mask}
  \end{subfigure}
  \begin{subfigure}{0.20\linewidth}
    \includegraphics[trim=0px 40px 0px 0px,clip,width=\linewidth]{figs/supp-relative-difference-pcons/cm}
    \caption{Camera motion}
  \end{subfigure}%
  \begin{subfigure}{0.20\linewidth}
    \includegraphics[trim=0px 40px 0px 0px,clip,width=\linewidth]{figs/supp-relative-difference-pcons/bsm}
    \caption{BG scene motion}
  \end{subfigure}%
  \begin{subfigure}{0.20\linewidth}
    \includegraphics[trim=0px 40px 0px 0px,clip,width=\linewidth]{figs/supp-relative-difference-pcons/fgd}
    \caption{FG displacement}
  \end{subfigure}%
  \begin{subfigure}{0.20\linewidth}
    \includegraphics[trim=0px 40px 0px 0px,clip,width=\linewidth]{figs/supp-relative-difference-pcons/fgm}
    \caption{FG pose motion}
  \end{subfigure}%
  \begin{subfigure}{0.20\linewidth}
    \includegraphics[trim=0px 40px 0px 0px,clip,width=\linewidth]{figs/supp-relative-difference-pcons/fgs}
    \caption{FG size}
  \end{subfigure}
  \caption{Relative improvement in temporal consistency when DEVIL attributes change from low to high.}
  \label{fig:supp-relative-difference-pcons}
\end{figure}

Figure~\ref{fig:supp-relative-difference-pcons} shows the relative change in temporal consistency performance (PCons) when each DEVIL attribute changes from low to high. Overall, we found that temporal consistency is the aspect of inpainting quality that is least sensitive to changes in DEVIL attributes; however, some models still experience more noticeable differences than others (\eg, DFCNet is remarkably sensitive to BG scene motion).

\newpage
\twocolumn

{\small
\bibliographystyle{ieee_fullname}
\bibliography{references}

\begin{thebibliography}{10}\itemsep=-1pt

\bibitem{noauthor_production_2018}
Production {Notes}: {Match} {Moving} {\textbar} {Nevada} {Film} {Office}, Feb.
  2018.
\newblock Library Catalog: nevadafilm.com.

\bibitem{barnes_patchmatch_2009}
Connelly Barnes, Eli Shechtman, Adam Finkelstein, and Dan~B Goldman.
\newblock {PatchMatch}: {A} {Randomized} {Correspondence} {Algorithm} for
  {Structural} {Image} {Editing}.
\newblock {\em ACM Transactions on Graphics}, 28(3):24:1--24:11, July 2009.

\bibitem{bay_surf_2006}
Herbert Bay, Tinne Tuytelaars, and Luc Van~Gool.
\newblock {SURF}: {Speeded} {Up} {Robust} {Features}.
\newblock In Aleš Leonardis, Horst Bischof, and Axel Pinz, editors, {\em
  European {Conference} on {Computer} {Vision}}, Lecture {Notes} in {Computer}
  {Science}, pages 404--417, Berlin, Heidelberg, 2006. Springer.

\bibitem{carreira_quo_2017}
Joao Carreira and Andrew Zisserman.
\newblock Quo {Vadis}, {Action} {Recognition}? {A} {New} {Model} and the
  {Kinetics} {Dataset}.
\newblock pages 6299--6308, 2017.

\bibitem{chang_free-form_2019}
Ya-Liang Chang, Zhe~Yu Liu, Kuan-Ying Lee, and Winston Hsu.
\newblock Free-{Form} {Video} {Inpainting} {With} {3D} {Gated} {Convolution}
  and {Temporal} {PatchGAN}.
\newblock In {\em {IEEE} {International} {Conference} on {Computer} {Vision}},
  Oct. 2019.

\bibitem{chang_learnable_2019}
Ya-Liang Chang, Zhe~Yu Liu, Kuan-Ying Lee, and Winston Hsu.
\newblock Learnable {Gated} {Temporal} {Shift} {Module} for {Deep} {Video}
  {Inpainting}.
\newblock In {\em British {Machine} {Vision} {Conference}}, Sept. 2019.

\bibitem{dollar_pedestrian_2012}
P. Dollar, C. Wojek, B. Schiele, and P. Perona.
\newblock Pedestrian {Detection}: {An} {Evaluation} of the {State} of the
  {Art}.
\newblock {\em IEEE Transactions on Pattern Analysis and Machine Intelligence},
  34(4):743--761, 2012.

\bibitem{ebdelli_video_2015}
Mounira Ebdelli, Olivier Le~Meur, and Christine Guillemot.
\newblock Video {Inpainting} with {Short}-{Term} {Windows}: {Application} to
  {Object} {Removal} and {Error} {Concealment}.
\newblock {\em IEEE Transactions on Image Processing}, 24(10):3034--3047, Oct.
  2015.
\newblock Conference Name: IEEE Transactions on Image Processing.

\bibitem{fischler_random_1981}
Martin~A. Fischler and Robert~C. Bolles.
\newblock Random {Sample} {Consensus}: {A} {Paradigm} for {Model} {Fitting}
  with {Applications} to {Image} {Analysis} and {Automated} {Cartography}.
\newblock {\em Communications of the ACM}, 24(6):381--395, June 1981.

\bibitem{forsyth_finding_1996}
David~A. Forsyth, Jitendra Malik, Margaret~M. Fleck, Hayit Greenspan, Thomas
  Leung, Serge Belongie, Chad Carson, and Chris Bregler.
\newblock Finding {Pictures} of {Objects} in {Large} {Collections} of {Images}.
\newblock In Jean Ponce, Andrew Zisserman, and Martial Hebert, editors, {\em
  International {Workshop} on {Object} {Representation} in {Computer}
  {Vision}}, Lecture {Notes} in {Computer} {Science}, pages 335--360, Berlin,
  Heidelberg, 1996. Springer.

\bibitem{gao_flow-edge_2020}
Chen Gao, Ayush Saraf, Jia-Bin Huang, and Johannes Kopf.
\newblock Flow-{Edge} {Guided} {Video} {Completion}.
\newblock In Andrea Vedaldi, Horst Bischof, Thomas Brox, and Jan-Michael Frahm,
  editors, {\em European {Conference} on {Computer} {Vision}}, Lecture {Notes}
  in {Computer} {Science}, pages 713--729, Cham, 2020. Springer International
  Publishing.

\bibitem{granados_background_2012}
Miguel Granados, Kwang~In Kim, James Tompkin, Jan Kautz, and Christian
  Theobalt.
\newblock Background {Inpainting} for {Videos} with {Dynamic} {Objects} and a
  {Free}-{Moving} {Camera}.
\newblock In {\em European {Conference} on {Computer} {Vision}}, pages
  682--695. Springer, 2012.

\bibitem{granados_how_2012}
M. Granados, J. Tompkin, K. Kim, O. Grau, J. Kautz, and C. Theobalt.
\newblock How {Not} to {Be} {Seen} — {Object} {Removal} from {Videos} of
  {Crowded} {Scenes}.
\newblock {\em Computer Graphics Forum}, 31(2pt1):219--228, 2012.

\bibitem{gupta_characterizing_2017}
Agrim Gupta, Justin Johnson, Alexandre Alahi, and Li Fei-Fei.
\newblock Characterizing and {Improving} {Stability} in {Neural} {Style}
  {Transfer}.
\newblock In {\em {IEEE} {International} {Conference} on {Computer} {Vision}},
  Oct. 2017.

\bibitem{he_mask_2017}
Kaiming He, Georgia Gkioxari, Piotr Dollar, and Ross Girshick.
\newblock Mask {R}-{CNN}.
\newblock In {\em {IEEE} {Conference} on {Computer} {Vision} and {Pattern}
  {Recognition}}, pages 2961--2969, 2017.

\bibitem{heusel_gans_2017}
Martin Heusel, Hubert Ramsauer, Thomas Unterthiner, Bernhard Nessler, and Sepp
  Hochreiter.
\newblock {GANs} {Trained} by a {Two} {Time}-{Scale} {Update} {Rule} {Converge}
  to a {Local} {Nash} {Equilibrium}.
\newblock In {\em Neural {Information} {Processing} {Systems}}, {NIPS}'17,
  pages 6629--6640, Long Beach, California, USA, Dec. 2017. Curran Associates
  Inc.

\bibitem{huang_temporally_2016}
Jia-Bin Huang, Sing~Bing Kang, Narendra Ahuja, and Johannes Kopf.
\newblock Temporally {Coherent} {Completion} of {Dynamic} {Video}.
\newblock {\em ACM Transactions on Graphics}, 35(6):196, 2016.

\bibitem{smugmug_inc_flickr_nodate}
SmugMug Inc.
\newblock Flickr.
\newblock https://www.flickr.com/.

\bibitem{kim_deep_2019}
Dahun Kim, Sanghyun Woo, Joon-Young Lee, and In~So Kweon.
\newblock Deep {Video} {Inpainting}.
\newblock In {\em {IEEE} {Conference} on {Computer} {Vision} and {Pattern}
  {Recognition}}, pages 5792--5801, 2019.

\bibitem{krizhevsky_imagenet_2012}
Alex Krizhevsky, Ilya Sutskever, and Geoffrey~E. Hinton.
\newblock {ImageNet} {Classification} with {Deep} {Convolutional} {Neural}
  {Networks}.
\newblock In {\em Neural {Information} {Processing} {Systems}}, 2012.

\bibitem{lee_copy-and-paste_2019}
Sungho Lee, Seoung~Wug Oh, DaeYeun Won, and Seon~Joo Kim.
\newblock Copy-and-{Paste} {Networks} for {Deep} {Video} {Inpainting}.
\newblock In {\em {IEEE} {International} {Conference} on {Computer} {Vision}},
  pages 4413--4421, 2019.

\bibitem{lin_microsoft_2014}
Tsung-Yi Lin, Michael Maire, Serge Belongie, James Hays, Pietro Perona, Deva
  Ramanan, Piotr Dollár, and C.~Lawrence Zitnick.
\newblock Microsoft {COCO}: {Common} {Objects} in {Context}.
\newblock In {\em European {Conference} on {Computer} {Vision}}, Lecture
  {Notes} in {Computer} {Science}, pages 740--755, Cham, 2014. Springer
  International Publishing.

\bibitem{newson_video_2014}
Alasdair Newson, Andrés Almansa, Matthieu Fradet, Yann Gousseau, and Patrick
  Pérez.
\newblock Video {Inpainting} of {Complex} {Scenes}.
\newblock {\em SIAM Journal on Imaging Sciences}, 7(4):1993--2019, 2014.

\bibitem{oh_onion-peel_2019}
Seoung~Wug Oh, Sungho Lee, Joon-Young Lee, and Seon~Joo Kim.
\newblock Onion-{Peel} {Networks} for {Deep} {Video} {Completion}.
\newblock In {\em {IEEE} {International} {Conference} on {Computer} {Vision}},
  pages 4403--4412, 2019.

\bibitem{perazzi_benchmark_2016}
Federico Perazzi, Jordi Pont-Tuset, Brian McWilliams, Luc Van~Gool, Markus
  Gross, and Alexander Sorkine-Hornung.
\newblock A {Benchmark} {Dataset} and {Evaluation} {Methodology} for {Video}
  {Object} {Segmentation}.
\newblock In {\em {IEEE} {Conference} on {Computer} {Vision} and {Pattern}
  {Recognition}}, pages 724--732, 2016.

\bibitem{rossler_faceforensics_2018}
Andreas Rössler, Davide Cozzolino, Luisa Verdoliva, Christian Riess, Justus
  Thies, and Matthias Nießner.
\newblock {FaceForensics}: {A} {Large}-{Scale} {Video} {Dataset} for {Forgery}
  {Detection} in {Human} {Faces}.
\newblock {\em arXiv}, 2018.

\bibitem{shen_first_2015}
Jie Shen, Stefanos Zafeiriou, Grigoris~G. Chrysos, Jean Kossaifi, Georgios
  Tzimiropoulos, and Maja Pantic.
\newblock The {First} {Facial} {Landmark} {Tracking} {In}-the-{Wild}
  {Challenge}: {Benchmark} and {Results}.
\newblock In {\em {IEEE} {International} {Conference} on {Computer} {Vision}
  {Workshops}}, Dec. 2015.

\bibitem{szegedy_going_2015}
Christian Szegedy, Wei Liu, Yangqing Jia, Pierre Sermanet, Scott Reed, Dragomir
  Anguelov, Dumitru Erhan, Vincent Vanhoucke, and Andrew Rabinovich.
\newblock Going {Deeper} {With} {Convolutions}.
\newblock pages 1--9, 2015.

\bibitem{vaswani_attention_2017}
Ashish Vaswani, Noam Shazeer, Niki Parmar, Jakob Uszkoreit, Llion Jones,
  Aidan~N. Gomez, Lukasz Kaiser, and Illia Polosukhin.
\newblock Attention is {All} {You} {Need}.
\newblock In {\em Neural {Information} {Processing} {Systems}}, 2017.

\bibitem{wang_video_2019}
Chuan Wang, Haibin Huang, Xiaoguang Han, and Jue Wang.
\newblock Video {Inpainting} by {Jointly} {Learning} {Temporal} {Structure} and
  {Spatial} {Details}.
\newblock In {\em {AAAI} {Conference} on {Artificial} {Intelligence}},
  volume~33, pages 5232--5239, 2019.

\bibitem{wexler_space-time_2007}
Yonatan Wexler, Eli Shechtman, and Michal Irani.
\newblock Space-{Time} {Completion} of {Video}.
\newblock {\em IEEE Transactions on Pattern Analysis and Machine Intelligence},
  29(3):463--476, 2007.
\newblock Publisher: IEEE.

\bibitem{wu_detectron2_2019}
Yuxin Wu, Alexander Kirillov, Francisco Massa, Wan-Yen Lo, and Ross Girshick.
\newblock Detectron2, 2019.
\newblock https://github.com/facebookresearch/detectron2.

\bibitem{xu_youtube-vos_2018}
Ning Xu, Linjie Yang, Yuchen Fan, Jianchao Yang, Dingcheng Yue, Yuchen Liang,
  Brian Price, Scott Cohen, and Thomas Huang.
\newblock {YouTube}-{VOS}: {Sequence}-to-{Sequence} {Video} {Object}
  {Segmentation}.
\newblock In Vittorio Ferrari, Martial Hebert, Cristian Sminchisescu, and Yair
  Weiss, editors, {\em European {Conference} on {Computer} {Vision}}, Lecture
  {Notes} in {Computer} {Science}, pages 603--619, Cham, 2018. Springer
  International Publishing.

\bibitem{xu_deep_2019}
Rui Xu, Xiaoxiao Li, Bolei Zhou, and Chen~Change Loy.
\newblock Deep {Flow}-{Guided} {Video} {Inpainting}.
\newblock In {\em {IEEE} {Conference} on {Computer} {Vision} and {Pattern}
  {Recognition}}, pages 3723--3732, 2019.

\bibitem{yeager_everything_2019}
Charles Yeager.
\newblock Everything {You} {Need} to {Know} {About} {Chroma} {Key} and {Green}
  {Screen} {Footage}, July 2019.
\newblock Library Catalog: www.premiumbeat.com Section: Video Production.

\bibitem{zeng_learning_2020}
Yanhong Zeng, Jianlong Fu, and Hongyang Chao.
\newblock Learning {Joint} {Spatial}-{Temporal} {Transformations} for {Video}
  {Inpainting}.
\newblock In Andrea Vedaldi, Horst Bischof, Thomas Brox, and Jan-Michael Frahm,
  editors, {\em European {Conference} on {Computer} {Vision}}, Lecture {Notes}
  in {Computer} {Science}, pages 528--543, Cham, 2020. Springer International
  Publishing.

\bibitem{zhang_internal_2019}
Haotian Zhang, Long Mai, Ning Xu, Zhaowen Wang, John Collomosse, and Hailin
  Jin.
\newblock An {Internal} {Learning} {Approach} to {Video} {Inpainting}.
\newblock In {\em {IEEE} {International} {Conference} on {Computer} {Vision}},
  Oct. 2019.

\bibitem{zhang_unreasonable_2018}
Richard Zhang, Phillip Isola, Alexei~A Efros, Eli Shechtman, and Oliver Wang.
\newblock The {Unreasonable} {Effectiveness} of {Deep} {Features} as a
  {Perceptual} {Metric}.
\newblock In {\em {IEEE} {Conference} on {Computer} {Vision} and {Pattern}
  {Recognition}}, pages 586--595, 2018.

\end{thebibliography}
}

\end{document}